\documentclass[letterpaper, 10 pt, conference]{ieeeconf}  

\IEEEoverridecommandlockouts                              

\usepackage{epsfig} 
\usepackage{mathptmx} 
\usepackage{times} 
\usepackage{amsmath} 
\usepackage{amssymb}  

\usepackage{xcolor}
\usepackage{array}
\usepackage[caption=false,font=normalsize,labelfont=sf,textfont=sf]{subfig}
\usepackage{textcomp}
\usepackage{stfloats}
\usepackage{url}
\usepackage{verbatim}
\usepackage{caption}
\usepackage{graphicx}
\usepackage{cite}
\usepackage{multirow}
\usepackage{booktabs}
\usepackage{xcolor}
\usepackage[x11names,table]{xcolor}
\usepackage{colortbl}
\definecolor{gr}{HTML}{E3EDD9}
\definecolor{bg}{HTML}{F0F0F0}

\title{\LARGE \bf
SurgRAW: Multi-Agent Workflow with Chain of Thought Reasoning for Robotic Surgical Video Analysis
}

\author{
Chang Han Low$^{1}$,
Ziyue Wang$^{1}$,
Tianyi Zhang$^{1,2}$,
Zhu Zhuo$^{1}$,
Zhitao Zeng$^{1}$,
Evangelos B. Mazomenos$^{3}$,\\
and Yueming Jin$^{1}$
\thanks{*This work was supported by Ministry of Education Tier 2 grant, Singapore (T2EP20224-0028), and Ministry of Education Tier 1 grant, Singapore (23-0651-P0001). Corresponding author: Yueming Jin.}%
\thanks{$^{1}$C. H. Low, Z. Wang, T. Zhang, Z. Zhu, Z. Zeng, and Y. Jin are with the National University of Singapore, Singapore
        {\tt\small ymjin@nus.edu.sg}}%
\thanks{$^{2}$T. Zhang is with the Bioinformatics Institute (BII), Agency for Science, Technology and Research (A*STAR), Singapore.}%
\thanks{$^{3}$E. B. Mazomenos is with the Wellcome/EPSRC Centre for Interventional and Surgical Sciences (WEISS) and the Department of Medical Physics and Biomedical Engineering, University College London, UK.}%
}

\begin{document}

\maketitle
\thispagestyle{empty}
\pagestyle{empty}

\begin{abstract}

Robotic-assisted surgery (RAS) is central to modern surgery, driving the need for intelligent systems with accurate scene understanding. Most existing surgical AI methods rely on isolated, task-specific models, leading to fragmented pipelines with limited interpretability and no unified understanding of RAS scene. Vision-Language Models (VLMs) offer strong zero-shot reasoning, but struggle with hallucinations, domain gaps and weak task-interdependency modeling. To address the lack of unified data for RAS scene understanding, we introduce \textbf{SurgCoTBench}, the first reasoning-focused benchmark in RAS, covering 14256 QA pairs with frame-level annotations across five major surgical tasks. Building on SurgCoTBench, we propose \textbf{SurgRAW}, a clinically aligned Chain-of-Thought (CoT) driven agentic workflow for zero-shot multi-task reasoning in surgery. SurgRAW employs a hierarchical reasoning workflow where an orchestrator divides surgical scene understanding into two reasoning streams and directs specialized agents to generate task-level reasoning, while higher-level agents capture workflow interdependencies or ground output clinically. Specifically, we propose a panel discussion mechanism to ensure task-specific agents collaborate synergistically and leverage on task interdependencies. Similarly, we incorporate a retrieval-augmented generation module to enrich agents with surgical knowledge and alleviate domain gaps in general VLMs. We design task-specific CoT prompts grounded in surgical domain to ensure clinically aligned reasoning, reduce hallucinations and enhance interpretability. Extensive experiments show that SurgRAW surpasses mainstream VLMs and agentic systems and outperforms a supervised model by 14.61\% accuracy. 
Dataset and code is available at https://github.com/jinlab-imvr/SurgRAW.git .

\end{abstract}

\section{Introduction}

Robotic surgical systems like da Vinci \cite{davinci} have driven the rapid adoption of robot-assisted surgery (RAS) in modern operating rooms, offering surgeons enhanced precision, dexterity, and access to anatomically challenging regions. Despite these advantages, the effectiveness of RAS ultimately depends on the surgeon’s ability to accurately interpret complex intraoperative scenes. Thus, the next frontier for advancing RAS lies in reliable, interpretable scene understanding that can provide meaningful cognitive assistance to surgeons. However, achieving robust surgical scene understanding remains challenging. Robotic surgical videos exhibit high visual complexity such as instruments and anatomical structures frequently overlapping or occluding one another, and their spatial relationships change rapidly across fine-grained surgical actions. These characteristics make intraoperative interpretation demanding even for experienced surgeons and underscore the need for computer-assisted approaches that can assist with interpretable surgical scene understanding.

Motivated by this need, AI-driven approaches \cite{jin2017sv,jin2020multi,twinanda2016endonet} have been developed to tackle surgical vision tasks such as workflow recognition, instrument presence detection, and error detection \cite{surgicalvqa,chen2024llm,gao2021trans,psychogyios2023sar,jin2017sv,yuan2025procedure,low2025cares}. However, these methods employ end-to-end networks to generate results without providing further explanations, limiting interpretability. Yet the model architectures for different tasks vary significantly, necessitating careful design and task-specific data collection for training. Moreover, most existing surgical datasets\cite{grasp,psiava,psychogyios2023sar,zhang2025csap} only provide one-to-three-task labels within the same scene. Thereby resulting in fragmentation which hampers the development of a unified, explainable surgical intelligence framework for RAS. 

Large Language Models (LLMs) have recently gained attention for their strong interaction capabilities in the zero-shot setting\cite{ChatGPT4o}. Prompting techniques like Chain-of-Thought (CoT)\cite{wei2022chain,llavacot} can elicit structured reasoning in general-domain LLMs. Vision-Language Models (VLMs) extend these capabilities to visual inputs and are increasingly applied to medical domains. In principle, these models offer cross-task generalizability and interpretability that overcomes the limitations of task-specific approaches.
Yet despite their promise, VLMs remain difficult to deploy reliably in RAS, due to several persistent and domain-specific challenges:
\textbf{(i)} Existing VLMs are primarily trained on natural scenes, resulting in substantial domain gaps when applied to robotic surgery. Although domain-specific finetuning efforts such as LLaVA-Med~\cite{llavamed}, LLaVA-Surg\cite{llavasurg} and SurgVLM \cite{zeng2025surgvlm} which aim to reduce this gap, they require large amounts of high-quality labeled surgical data~\cite{padoy2019machine, twinanda2016endonet} which are scarce. 
\textbf{(ii)} VLMs are also prone to hallucinations and prompt sensitivity, undermining their reliability in safety-critical surgical settings. Prompt engineering methods like LLaVA-CoT~\cite{llavacot} attempt to mitigate these issues by automatically generating CoT for each query. However, these auto-generated explanations often exacerbate hallucinations rather than reduce them in the surgical domain, as the model lacks the domain understanding needed to reason accurately.
\textbf{(iii)} VLMs typically treat surgical tasks independently, overlooking the workflow dependencies and hierarchical reasoning structure inherent in surgery. In contrast, multi-agent LLM frameworks like MDAgents\cite{kim2024mdagents} and MedAgents \cite{tang2024medagents} demonstrate that coordinated, specialized agents can decompose complex problems and maintain contextual consistency. However, their potential for surgical scene understanding remains largely underexplored, in part because existing surgical datasets do not offer unified, multi-task workflow annotations that enables systematic evaluation of reasoning capabilities.

To address these issues, we propose a \textbf{R}easoning multi-\textbf{A}gent \textbf{W}orkflow for \textbf{Surg}ical intelligence, termed SurgRAW. Designed for RAS, SurgRAW implements a hierarchical reasoning framework involving a central orchestrator which divides surgical scene understanding tasks into two streams, namely Visual-Semantic (VS) and Cognitive-Inference (CI). Subsequently, the respective coordinators of the streams will further delegate the tasks to specialized VLM agents for task-level reasoning while higher level agents account for workflow-level dependencies or reinforce clinical grounding.
Specifically, we introduce a novel panel discussion mechanism for VS pathway to initiate multi-agent collaboration and debate, explicitly leveraging task interdependencies to refine collective reasoning. By verifying the consistency between responses and intermediate steps across different agents, SurgRAW facilitates rich interactions among agents, enabling a comprehensive analysis of robotic surgical videos.
For CI pathway, we integrate Retrieval-Augmented Generation (RAG) to bridge the domain gap, ensuring that decisions are grounded in clinically-aligned surgical guidelines.

Rather than fine-tuning VLMs or relying on LLM-generated CoTs, we designed task-specific CoT prompts tailored to key elements of RAS such as instruments, tissue and action recognition. These structured prompts mitigate hallucinations by dividing surgical tasks from classification into a series of conductable observation and analysis. By guiding step-by-step logical deductions, CoT enhances explainability and accuracy. Overall, our contribution can be summarized as below: 

\begin{itemize}
\item We introduce SurgCoTBench, the first reasoning-focused benchmark for RAS, unifying five major vision tasks within unified robotic surgical scenarios for systematic evaluation.
\item We propose SurgRAW, one-of-the-first agentic frameworks for RAS capable of addressing key tasks across the surgical workflow with structured explainability.
\item We introduce a panel discussion mechanism and an RAG module to enable agents to share insights and generate reliable responses.
\item We equip VLM agents with task-specific CoT prompts designed to enhance transparent, step-by-step reasoning that is consistent with RAS workflow structure.
\end{itemize}
\section{Related Work}

\subsection{Surgical Scene Understanding}
Surgical scene understanding has evolved through task-specific solutions including foundation model-based segmentation approaches like SurgicalSAM~\cite{yue2024surgicalsam}, spatiotemporal networks for action recognition, and temporal convolutional networks for phase classification~\cite{surgicalvqa,chen2024llm,gao2021trans,jin2017sv,cog}. However, these isolated approaches differ significantly in architecture and design principles while overlooking intrinsic correlations among surgical tasks, hindering real-world deployment where multiple tasks should be addressed simultaneously. 

\subsection{LLM-based AI Agents}
LLM-based agents have demonstrated remarkable reasoning and decision-making capabilities across diverse applications including industrial engineering~\cite{a1} and embodied agents~\cite{a7,a9}. Recent medical LLM-based agentic frameworks\cite{tang2024medagents,kim2024mdagents,wang2025medagent} further highlight the potential of multi-agent LLM systems in medical settings, showcasing improved clinical reasoning and coordinated decision-making. Despite these advancements, such agent-based approaches remain largely unexplored in surgical domains. The inherent reasoning capabilities of LLMs demonstrates the potential for developing intelligent surgical systems that can provide contextual guidance and enhance surgical outcomes.

\subsection{Medical and Surgical VLMs}
Recent advances in medical and surgical VLMs highlight growing interest in multimodal clinical reasoning. General-purpose medical VLMs such as Med-Gemma\cite{sellergren2025medgemma} and LLaVA-Med\cite{llavamed} aim to extend open-source architectures to radiology and pathology image interpretation. Similarly, Surgical VLMs\cite{wang2025endochat,zeng2025surgvlm,li2024llava,li2025surgpub, wang2025surgvidlm} demonstrate promising visual–language understanding but remain focused on narrow perceptual tasks. However, existing surgical VLMs require large-scale training data and substantial computational resources, yet still lack extensive reasoning supervision.

\begin{figure*}[htbp]
\centering
\includegraphics[width=\textwidth]{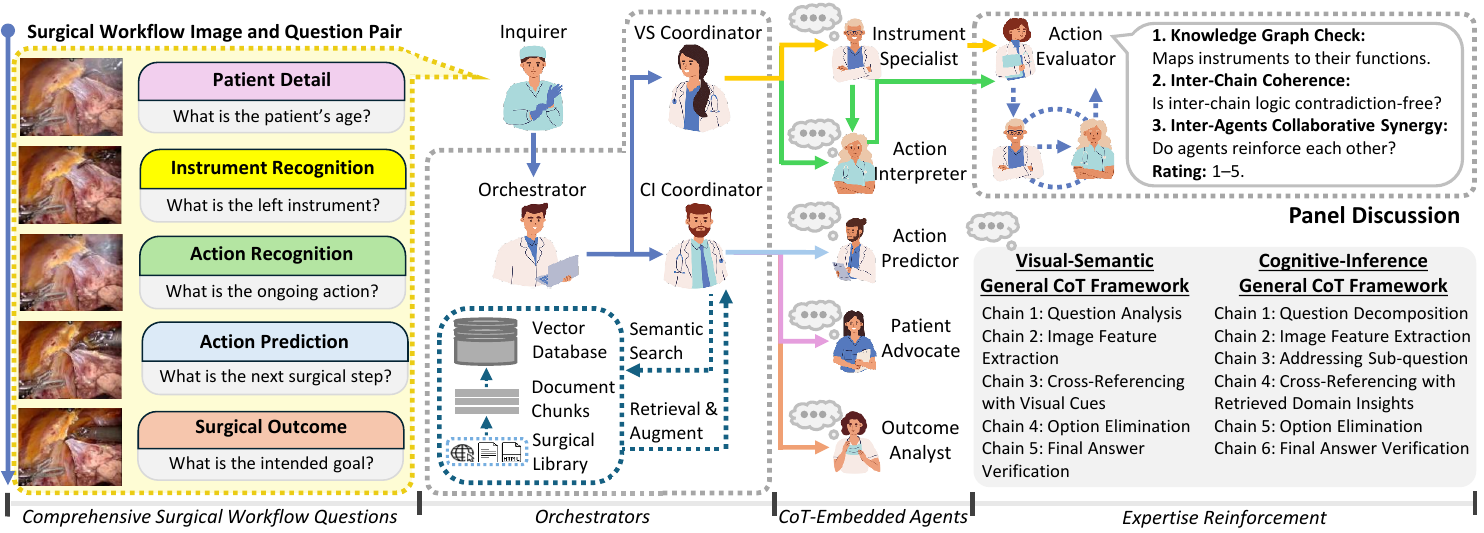}
\caption{The overall structure of SurgRAW processes surgical queries through hierarchical orchestrators and CoT-embedded expert agents, with RAG and panel discussions enhancing accuracy and domain reliability.} 
\label{fig:main}
\vspace{-10pt}
\end{figure*}

\section{Methodology}

\subsection{SurgCOTBench Dataset}
Existing surgical video datasets \cite{psiava,grasp,psychogyios2023sar,twinanda2016endonet,zhang2025csap} are fragmented in scope, typically annotating only one to three vision tasks such as instrument or action recognition, and none capturing the multi-step reasoning processes essential for surgical decision-making. As a result, they are insufficient for evaluating holistic surgical understanding or developing models capable of generalizing across different surgical tasks. To address these gaps, we introduce \textbf{SurgCoTBench}, a frame-level, multi-task reasoning benchmark for robotic-assisted surgery. The dataset is constructed from 12 patient videos across prostatectomy and lobectomy procedures. In total, SurgCoTBench contains \textbf{2,277 curated frames} and \textbf{14,256 vision-language Question-Answer (QA) pairs} spanning five core reasoning tasks, as shown in Table~\ref{tab:merged_dataset_table}. Following established data collection practices~\cite{yuan2023learning,schmidgall2024general,llavasurg}, surgical recordings were sourced from publicly available da Vinci prostatectomy and lobectomy videos. Transcripts were extracted using WhisperX~\cite{radford2023robust}, and videos were downsampled to 1\, frames per second (FPS). Frames were selected to capture clinically meaningful phases such as bladder neck dissection in prostatectomy and lymph-node dissection in lobectomy. For supervised models, the dataset is split by patient, with 8 videos for training and 4 videos for testing.

\begin{table}[htbp]
\centering
\caption{Distribution of SurgCoTBench}
\begin{tabular}{lccrr}
\hline
\multicolumn{5}{c}{\textit{\textbf{Prostatectomy}}} \\
\hline
\textbf{Category}           & \textbf{\#Frames} & \textbf{\#QA Pairs} & \textbf{\#Left} & \textbf{\#Right} \\
\hline
Action Recognition     & 1,542 & 2,177 & 1,119 & 1,058 \\
Instrument Recognition & 1,542 & 2,469 & 1,122 & 1,347 \\
Action Prediction      & 1,542 & 1,622 & --    & --    \\
Surgical Outcome       & 1,542 & 1,542 & --    & --    \\
Patient Detail         & 1,542 & 1,542 & --    & --    \\
\textbf{Procedure Total} & \textbf{1,542} & \textbf{9,352} & -- & -- \\
\hline
\multicolumn{5}{c}{\textit{\textbf{Lobectomy}}} \\
\hline
\textbf{Category}           & \textbf{\#Frames} & \textbf{\#QA Pairs} & \textbf{\#Left} & \textbf{\#Right} \\
\hline
Action Recognition     & 735 & 1,331 & 637 & 694 \\
Instrument Recognition & 735 & 1,368 & 640 & 728 \\
Action Prediction      & 735 & 735   & --  & --  \\
Surgical Outcome       & 735 & 735   & --  & --  \\
Patient Detail         & 735 & 735   & --  & --  \\
\textbf{Procedure Total} & \textbf{735} & \textbf{4,904} & -- & -- \\
\hline
\addlinespace[2pt]
\textbf{Overall}            & \textbf{2,277} & \textbf{14,256} & -- & -- \\
\hline
\end{tabular}
\label{tab:merged_dataset_table}
\vspace{-10pt}
\end{table}

\subsubsection{Reasoning Taxonomy in SurgCoTBench}
We develop a structured taxonomy of five surgical reasoning tasks including instrument recognition, action recognition, action prediction, patient-detail extraction, and outcome assessment, shown in Table \ref{tab:task_categories}. This reflects the hierarchy of perceptual, spatial, and semantic decisions routinely made during robotic surgery. Instrument and action recognition tasks are further disaggregated into left- and right-arm perspectives to capture fine-grained bimanual reasoning. This taxonomy ensures clinical validity and comprehensive coverage of intraoperative reasoning demands.

\setlength{\tabcolsep}{2mm}
\begin{table}[htbp]
\centering
\caption{SurgCoTBench reasoning task categories.}
\begin{tabular}{p{1.8cm}p{6.1cm}}
\hline
\textbf{Task Category} & \textbf{Description} \\
\hline

Action Recognition & Determining the specific maneuver performed by each instrument, structured for both left- and right-arm perspectives. \\

Instrument Recognition & Identifying surgical instruments attached to robotic arms based on visual appearance, tool geometry, and contextual cues. \\

Action Prediction & Anticipating the next surgical action from current tool configuration, anatomy exposure, and procedural context, evaluating temporal and planning reasoning. \\

Patient Detail & Inferring demographic and clinical attributes (e.g., age group) from anatomical appearance, organ characteristics, and visual context. \\

Surgical Outcome & Identifying the primary surgical outcome visible in the frame based on tool–tissue interactions, anatomical state, and procedural cues. \\
\hline
\end{tabular}
\label{tab:task_categories}
\vspace{-10pt}
\end{table}

\subsubsection{SurgCoTBench Construction}
For each selected frame, GPT-4o~\cite{ChatGPT4o} generates structured multiple-choice questions across all task types, producing up to seven QA pairs per frame. Notably, the surgical YouTube videos featured detailed real-time verbal explanations from the operating surgeon. When transcribed using Whisper\cite{radford2023robust}, these narrations yield content-rich, clinically accurate captions that provides semantic grounding well beyond the raw visual stream, enabling high-quality question generation. The constructed data shows high reliability.
First, the narrations originate from operating surgeons, therefore the transcription naturally reflects clinically correct terminology and procedural interpretation, providing a trustworthy foundation for question generation. Second, SurgCoTBench undergoes manual review by two clinicians who jointly verify semantic correctness, instrument terminology, alignment with the visual scene, and consistency across procedures, establishing consensus through collaborative review. Finally, the five reasoning tasks in SurgCoTBench are aligned with mainstream surgical AI tasks, while also extending the scope of existing benchmarks. In particular, \textit{Patient Detail} and \textit{Surgical Outcome} represent forms of implicit contextual reasoning, as the required information is not directly visible in the frame. The model must infer using indirect visual cues such as tissue, instrument, and anatomy.
These tasks extend the benchmark beyond explicit recognition toward nuanced, context-dependent reasoning that reflects higher-level intraoperative understanding.

\subsection{Surgical Multi-Agent Workflow}

\textbf{\begin{figure*}[thbp]
\centering
\includegraphics[width=0.95\textwidth]{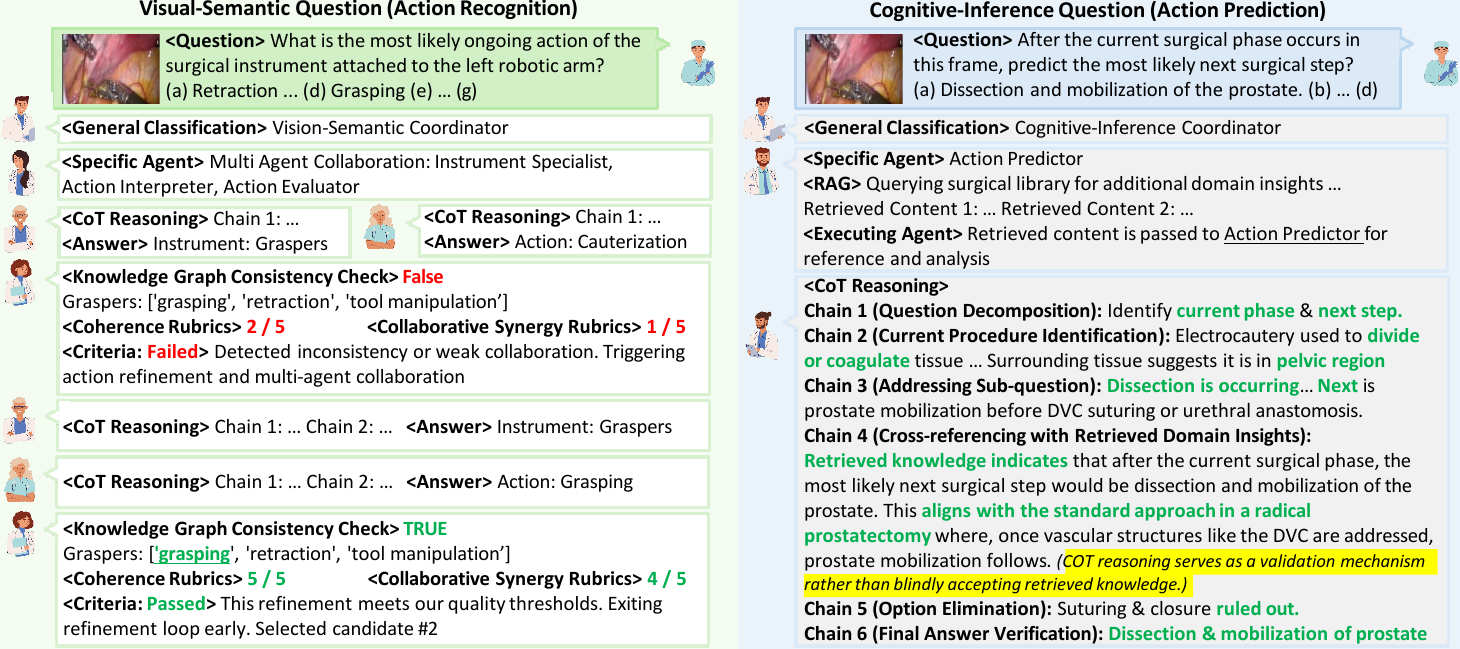}
\caption{Chat board demonstrating SurgRAW's overall workflow and intermediate reasoning for both reasoning streams.} 
\label{fig:example}
\vspace{-10pt}
\end{figure*}}

\vspace{-7pt}
In this work, we observe that surgical scene understanding involves two different streams of reasoning. The first stream consists of VS tasks, which focus on perceptual analysis such as identifying instruments, recognizing actions, and describing what is visible in the frame. The second stream consists of CI tasks which rely on understanding procedural logic, patient-related context or predicting surgical outcome from implicit cues. To holistically integrate and handle these two levels of reasoning tasks within a single unified framework, we propose a hierarchical multi-agent system named SurgRAW. As shown in Fig.~\ref{fig:main}, SurgRAW begins with a top-level orchestrator that classifies each incoming question as either a VS or a CI task. Based on this classification, the question is subsequently routed to one of two secondary orchestrators, the VS Coordinator or the CI Coordinator, which then assigns the question to the appropriate task-specific agent.

In this work, we focus on five key surgical vision tasks, each handled by a dedicated task-specific agent: the  Instrument Specialist, Action Interpreter, Action Predictor, Patient Advocate and Outcome Analyst. As VS and CI tasks have fundamentally different characteristics, they require different forms of reasoning mechanism. VS tasks often involve ambiguous visual cues. Hence, we add an auxiliary evaluator agent that conducts and moderates a panel discussion to cross-check the outputs of the VS agents. To reinforce accuracy in VS tasks, the evaluator uses a compact surgical knowledge graph (KG) that encodes valid instrument–action relationships. This serves as a lightweight constraint module that filters out unlikely predictions and maintains clinical plausibility. In contrast, CI tasks rely on contextual or procedural cues that cannot be inferred from visual information alone. For these tasks, SurgRAW integrates RAG to provide external surgical knowledge that helps the CI agents reason about patient details, outcomes, or procedural logic. The final response is produced after these intermediate steps.

To improve the transparency and reliability of each agent, we incorporate task-specific CoT prompts. Instead of relying on generic CoT generated by LLMs \cite{llavacot}, we design structured CoT tailored to the characteristics of each task. These CoT guide the agents to explain their reasoning step by step, reduce hallucinations, and make the decision process easier to verify. VS agents follow CoT focused on visual inspection and evidence extraction, while CI agents use CoT that emphasize contextual reasoning and clinical logic.

\subsection{Expertise Reinforcement through Supplementary Modules}
\textbf{Panel Discussion for Visual-Semantic Tasks.}
Despite acquiring strong reasoning abilities through CoT, multi-agent collaboration can further benefit task-level agents to achieve a more comprehensive analysis \cite{kim2025mdagents,li2024mmedagent}. 
VS tasks, in particular, demand contextual awareness and cross-verification to ensure accuracy. For instance, the Action Interpreter may consult the Instrument Specialist to confirm the identity or presence of a surgical instrument before proceeding. 

To tackle the challenge, SurgRAW introduces an Action Evaluator to conduct panel discussion from three perspectives:
i) To safeguard prediction-level consistency, the Action Evaluator integrates a surgical knowledge graph detailing permissible instrument-action relationships which is derived from the \textit{Da Vinci} surgical robot’s official specifications \cite{davinci}.  This encourages different task-level agents to cross-reference their predictions and check for procedural alignment. ii) To enhance consistency and quality at the reasoning level, the Action Evaluator employs two additional structured rubrics: the \textbf{Inter-Chain Coherence} rubric evaluates the logical alignment of CoT results within a task, ensuring a reliable and logical reasoning process.
Meanwhile, the \textbf{Inter-Agents Collaborative Synergy} rubric assesses how well agents reinforce each other’s predictions while mitigating error propagation, preventing early inconsistencies from affecting final decisions. 
During the panel discussion, the Action Evaluator pass the information between task-level agents to refine the response until the KG check reaches \textbf{true}, and the two rubrics are higher than their corresponding threshold of 3 out of 5. The conversation will be interrupted if max iteration of 3 runs is reached. This maximum iteration cap is imposed as a pragmatic safeguard to bound refinement and prevent unproductive repeated disagreement. These evaluations foster inter-agent synergy and reinforce evidence-based reasoning.

\noindent \textbf{RAG for Cognitive-Inference Tasks.}
To strengthen cognitive-inference decision-making, SurgRAW integrates a RAG module to improve reliability and relevance of responses \cite{gao2023retrieval,lewis2020retrieval}. The CI Coordinator queries repository populated with medical resources from MedlinePlus \cite{national2006medlineplus,miller2000medlineplus}.The retrieved content is dynamically incorporated into the corresponding CoT-embedded VLM agents, allowing them to refine outputs by cross-referencing validated medical information. This ensures alignment with established medical standards while reducing hallucinations. Through visual evidence validation and cross-referencing medically grounded retrieved documents, RAG improves SurgRAW’s dependability and makes AI-driven surgical assistance more transparent and clinically actionable.

\subsection{Chain-of-Thought Prompt Design for Surgery}
Surgical scene understanding requires contextual assessment, anatomical understanding, prior domain knowledge, and the ability to make sense of unclear visual cues \cite{hashimoto2018artificial,liu2025deep,manning2009cognitive,national2006medlineplus}. Our CoT prompting framework structures the model’s reasoning by guiding VLMs such as GPT-4o \cite{ChatGPT4o} through sequential analytical steps \cite{sivarajkumar2024empirical}.
Although each of the five task employs its own chain definitions, we identify distinct overarching reasoning structure for VS and CI categories, which govern the CoTs design of the tasks (Fig.~\ref{fig:main}).

\textbf{VS CoTs} (Action Recognition and Instrument Recognition) follow a unified five-stage reasoning structure. The process begins with question analysis, where the model interprets the intent and constraints of the query. This is followed by image feature extraction, in which relevant objects, spatial relations, and visual attributes are identified. The third stage cross-references these extracted features with the expectations implied by the question, ensuring alignment between visual evidence and task requirements. The CoT then proceeds to option elimination, removing options that are incompatible with the established visual-contextual cues. 
Finally, the answer is verified through a consistency check against the visual scene. 
The VS tasks adhere to this CoT generation process, while each specific task extends or specializes some stages. 

\subsubsection{\textbf{Action Recognition.}}
\texttt{Chain~1} analyzes the question to determine which surgical action is being queried and what visual or procedural cues are relevant. \texttt{Chain~2} extracts image features such as instrument motion, interaction point and spatial relationships that are indicative of ongoing actions.\texttt{Chain~3} cross-references these features with the action candidates, checking whether the observed instrument–tissue interactions match the expected patterns implied by the question. \texttt{Chain~4} eliminates action options that conflict with visible motion cues or violate typical workflow logic. \texttt{Chain~5} verifies the final choice by confirming its consistency with the overall scene context and current phase.

\subsubsection{\textbf{Instrument Recognition.}}
\texttt{Chain~1} identifies what instrument-related information the question is asking for. \texttt{Chain~2.1} localizes the visible parts of the instrument, even when partially occluded, and \texttt{Chain~2.2} extracts key structural cues such as surface texture, jaw and shaft geometry. \texttt{Chain~2.3} infers missing or occluded parts based on typical instrument designs, and \texttt{Chain~3.1} evaluates how the instrument interacts with tissue to narrow its functional category. \texttt{Chain~3.2} resolves ambiguities between functionally similar tools based on visual features. \texttt{Chain~3.3} cross-verifies these cues against expected functions within the phase. \texttt{Chain~4} eliminates conflicting candidates. \texttt{Chain~5} performs a final verification and selects the instrument that best matches all observed features and contextual evidence.


\textbf{CI CoTs} (Surgical Plan, Surgical Outcome, Patient Detail) follow a structured reasoning process that differs from VS tasks in how they begin. Unlike VS CoTs, which explicitly analyze the input questions and visible elements, CI tasks often involve information that is implicit or only partly visible. Therefore, CI CoTs start by decomposing the question into its clinical components. The chain then extracts the visual cues relevant to each component, addresses the resulting sub-questions, and refines these intermediate conclusions using procedural norms or retrieved surgical knowledge. Next, the process eliminates answer options that conflict with either the visual evidence or retrieved content. The reasoning concludes with a final verification step to ensure the selected answer is consistent with the overall surgical context.

\subsubsection{\textbf{Action Prediction.}}
\texttt{Chain~1} deconstructs the question into sub-questions about the current and the next expected phase. \texttt{Chain~2} identifies the current procedural phase by examining instrument–tissue interactions. 
\texttt{Chain~3} answers these sub-components using visual and contextual evidence. \texttt{Chain~4} refines these answers by referencing procedural standards or retrieved knowledge.

\subsubsection{\textbf{Surgical Outcome.}}
\texttt{Chain~1} splits the questions into granular sub-questions for a multi-perspective analysis. \texttt{Chain~2} extracts the relevant visual evidence by determining the ongoing action and the tissue involved. \texttt{Chain~3} addresses the sub-questions by combining these visual cues obtaining intermediate inferences. \texttt{Chain~4} cross-reference the intermediate inferences with retrieved knowledge. 

\subsubsection{\textbf{Patient Detail.}}
\texttt{Chain~1} decomposes the question by identifying the specific patient attributes being requested. \texttt{Chain~2} extracts relevant visual information. \texttt{Chain~3} addresses the decomposed sub-question by matching the extracted visual cues to the candidate options. \texttt{Chain~4} refines these matches by interpreting unclear or partial cues using retrieved content. 

Across all CI tasks, \texttt{Chain~5} removes options that conflict with procedural reasoning or retrieved knowledge, reducing hallucination. \texttt{Chain~6} synthesizes the final conclusion.

\section{Experiment}

\begin{table*}[!th]
\caption{Comparison results on SurgCoTBench (\% mean accuracy). Best results are bolded. Second-best results are underlined. All models are evaluated in a zero-shot setting, except supervised baseline which requires training.}
\begin{center}
\resizebox{\textwidth}{!}{%
\begin{tabular}{lcccccccc}
\toprule
\multirow{2}{*}{Methods} &
\multirow{2}{*}{Overall} &
\multicolumn{4}{c}{Cognitive-Inference Tasks} &
\multicolumn{3}{c}{Visual-Semantic Tasks} \\
\cmidrule(lr){3-6} \cmidrule(lr){7-9}
 & & Act. Pred. & Out. & Pat. & Avg. & Act. Rec. & Inst. Rec. & Avg. \\

\midrule
\rowcolor{bg}
\multicolumn{9}{c}{\bfseries \textit{Supervised Method 
(Training-Required)}} \\  \midrule

\addlinespace[2pt]
Surgical-VQA \cite{surgicalvqa} (Supervised) &
47.12$\pm$5.46 &
50.58$\pm$2.16 &
49.83$\pm$1.91 &
73.87$\pm$2.40 &
58.76$\pm$11.18 &
38.31$\pm$3.54 &
34.42$\pm$1.13 &
30.37$\pm$1.79 \\
 
\midrule
\rowcolor{bg}
\multicolumn{9}{c}{\bfseries \textit{Zero-Shot Setting}} \\  \midrule

\multicolumn{9}{l}{\textbf{General VLMs}} \\
\addlinespace[2pt]

GPT-4o \cite{ChatGPT4o} &
31.44$\pm$2.57 &
32.98$\pm$5.81 &
17.92$\pm$7.33 &
9.44$\pm$5.62 &
20.11$\pm$2.98 &
37.21$\pm$7.94 &
43.57$\pm$8.83 &
40.39$\pm$3.04 \\

Qwen3VL-8B \cite{yang2025qwen3} &
35.69$\pm$0.47 &
26.45$\pm$2.10 &
23.63$\pm$1.66 &
40.41$\pm$0.45 &
30.16$\pm$1.10 &
30.16$\pm$0.91 &
\underline{59.46$\pm$0.06} &
44.81$\pm$0.48 \\

LLava-OV-7B \cite{llava-1v} &
27.63$\pm$0.56 &
19.00$\pm$1.69 &
39.63$\pm$2.01 &
30.23$\pm$1.44 &
30.74$\pm$1.03 &
19.16$\pm$1.59 &
30.88$\pm$1.10 &
25.12$\pm$1.21 \\
\midrule
\addlinespace[2pt]

\multicolumn{9}{l}{\textbf{Domain-adapted Medical VLMs}} \\
\addlinespace[2pt]


MedGemma-4B \cite{sellergren2025medgemma} &
24.17$\pm$0.99 &
20.28$\pm$1.27 &
39.74$\pm$1.25 &
34.01$\pm$0.88 &
31.91$\pm$1.13 &
32.85$\pm$0.72 &
0.33$\pm$0.64 &
16.59$\pm$0.68 \\

LLaVA-Med-7B \cite{llavamed} &
12.46$\pm$0.13 &
15.39$\pm$1.47 &
16.69$\pm$1.32 &
21.94$\pm$0.07 &
18.01$\pm$2.83 &
8.44$\pm$2.43 &
7.42$\pm$4.32 &
7.93$\pm$0.51 \\

\midrule
\addlinespace[2pt]


\multicolumn{9}{l}{\textbf{Prompt Engineering Technique}} \\
\addlinespace[2pt]

LLaVa-CoT \cite{llavacot} (GPT-4o) &
38.92$\pm$1.83 &
34.11$\pm$2.48 &
23.94$\pm$2.97 &
62.18$\pm$3.91 &
40.74$\pm$2.02 &
35.62$\pm$2.71 &
33.48$\pm$3.05 &
34.55$\pm$1.81 \\

LLaVa-CoT \cite{llavacot} (Qwen3VL-8B) &
34.91$\pm$0.12 &
18.56$\pm$0.64 &
\underline{44.31$\pm$0.05} &
52.89$\pm$0.49 &
38.59$\pm$0.23 &
8.30$\pm$0.07 &
52.67$\pm$0.12 &
31.48$\pm$0.08 \\
\midrule
\addlinespace[2pt]

\multicolumn{9}{l}{\textbf{Agentic frameworks}} \\
\addlinespace[2pt]

MDAgents \cite{kim2024mdagents} (GPT-4o) &
32.88$\pm$2.31 &
34.27$\pm$5.92 &
19.84$\pm$7.51 &
7.02$\pm$5.11 &
20.38$\pm$2.72 &
39.14$\pm$7.88 &
46.92$\pm$9.13 &
41.77$\pm$3.01 \\

MedAgents \cite{tang2024medagents} (GPT-4o) &
31.96$\pm$2.85&
34.10$\pm$6.40&
18.35$\pm$7.90 &
7.88$\pm$5.92 &
20.11$\pm$3.21 &
38.02$\pm$8.65 &
45.40$\pm$9.72 &
41.71$\pm$3.44 \\

\midrule
\multicolumn{9}{l}{\textbf{Ours}} \\
\addlinespace[2pt]

SurgRAW-GPT4o &
\underline{61.73 $\pm$2.42} &
\textbf{70.48$\pm$2.81} &
42.87$\pm$3.11 &
\textbf{96.21$\pm$3.53} &
\textbf{70.39$\pm$2.16} &
\underline{44.52$\pm$2.88} &
54.13$\pm$1.93 &
\underline{49.81$\pm$2.31} \\

SurgRAW-Qwen3VL-8B  &
\textbf{64.03$\pm$1.51} &
\underline{51.65 $\pm$1.25} &
\textbf{50.82$\pm$5.33} &
\underline{86.12$\pm$2.84} &
\underline{62.86$\pm$2.06} &
\textbf{62.47$\pm$1.70} &
\textbf{67.33$\pm$3.92} &
\textbf{65.01$\pm$2.20} \\

\bottomrule
\end{tabular}}
\label{tab1}
\end{center}
\vspace{-10pt}
\end{table*}
\vspace{-5pt}

\subsection{Implementation and Evaluation Metrics}

We compare SurgRAW against five categories of state-of-the-art approaches: (1) \textit{general VLMs}, including GPT-4o \cite{ChatGPT4o} and Qwen3VL-8B \cite{yang2025qwen3}; (2) \textit{domain-adapted medical VLMs}, such as MedGemma \cite{sellergren2025medgemma} and LLaVA-Med \cite{llavamed}; (3) \textit{CoT-based prompting approaches}, including LLaVA-CoT \cite{llavacot}; (4) \textit{medical agentic reasoning frameworks}, including MDAgents \cite{kim2024mdagents} and MedAgents \cite{tang2024medagents}; and (5) a \textit{supervised baseline}, Surgical-VQA \cite{surgicalvqa}. All compared methods adhere strictly to their original setting. For all open-source VLMs, we set a maximum token limit of 2048, top-k of 40, and both top-p and temperature set to 0.8 to ensure fair comparison. For SurgicalVQA \cite{surgicalvqa}, we follow the original training settings. All experiments on VLMs were conducted in a zero-shot setting. SurgRAW incurs additional inference overhead compared to a single-pass VLM due to multi-agent system.
Model inferences for open-sourced VLMs are performed on a single NVIDIA A6000 GPU. Evaluation is based on accuracy rate (Acc) and all reported standard deviations are computed over three independent runs.

\subsection{Comparison with State-of-the-Art Methods}

We evaluate SurgRAW against aforementioned four-category models:  general VLMs, domain-adapted medical VLMs, CoT-based approaches, and agentic reasoning frameworks (Table~\ref{tab1}). General VLMs perform moderately on VS tasks but struggle with CI due to missing surgical priors. Domain-adapted medical VLMs do not close this gap and often underperform general VLMs, showing that broad medical adaptation is insufficient for surgical reasoning. LLaVA-CoT \cite{llavacot} yields limited gains through structured prompting, and existing agentic frameworks in the medical domain\cite{kim2024mdagents,tang2024medagents} still transfer poorly to real surgical workflows. 
In contrast, SurgRAW yields substantial improvements \textbf{regardless on closed-sourced VLM \cite{ChatGPT4o} or open-sourced VLM\cite{yang2025qwen3}}. 
Across matched VLM backbones, SurgRAW yields an average overall improvement of \textbf{29.32\%} over standard VLMs. Performance gains are most pronounced for CI task where SurgRAW-GPT4o achieves a \textbf{37.50\%} increase in Action Prediction and a remarkable \textbf{86.77\%} boost in Patient Detail Extraction compared to its baseline GPT4o. Regardless of VLM backbone in a zeroshot setting, SurgRAW surpasses the supervised Surgical-VQA~\cite{surgicalvqa} across all metrics. 
Crucially, SurgRAW demonstrates remarkable reliability. Unlike Surgical-VQA \cite{surgicalvqa}, a supervised baseline which exhibits high standard deviation ($\pm$5.46\%), our method maintains consistent performance ($\pm$\textbf{1.51}\%) even while achieving state-of-the-art accuracy
We also show the qualitative examples in Fig.~\ref{fig:example}, where SurgRAW generates coherent and clinically aligned reasoning traces.

\begin{figure*}[htbp]
\centering
\includegraphics[width=0.85\textwidth]{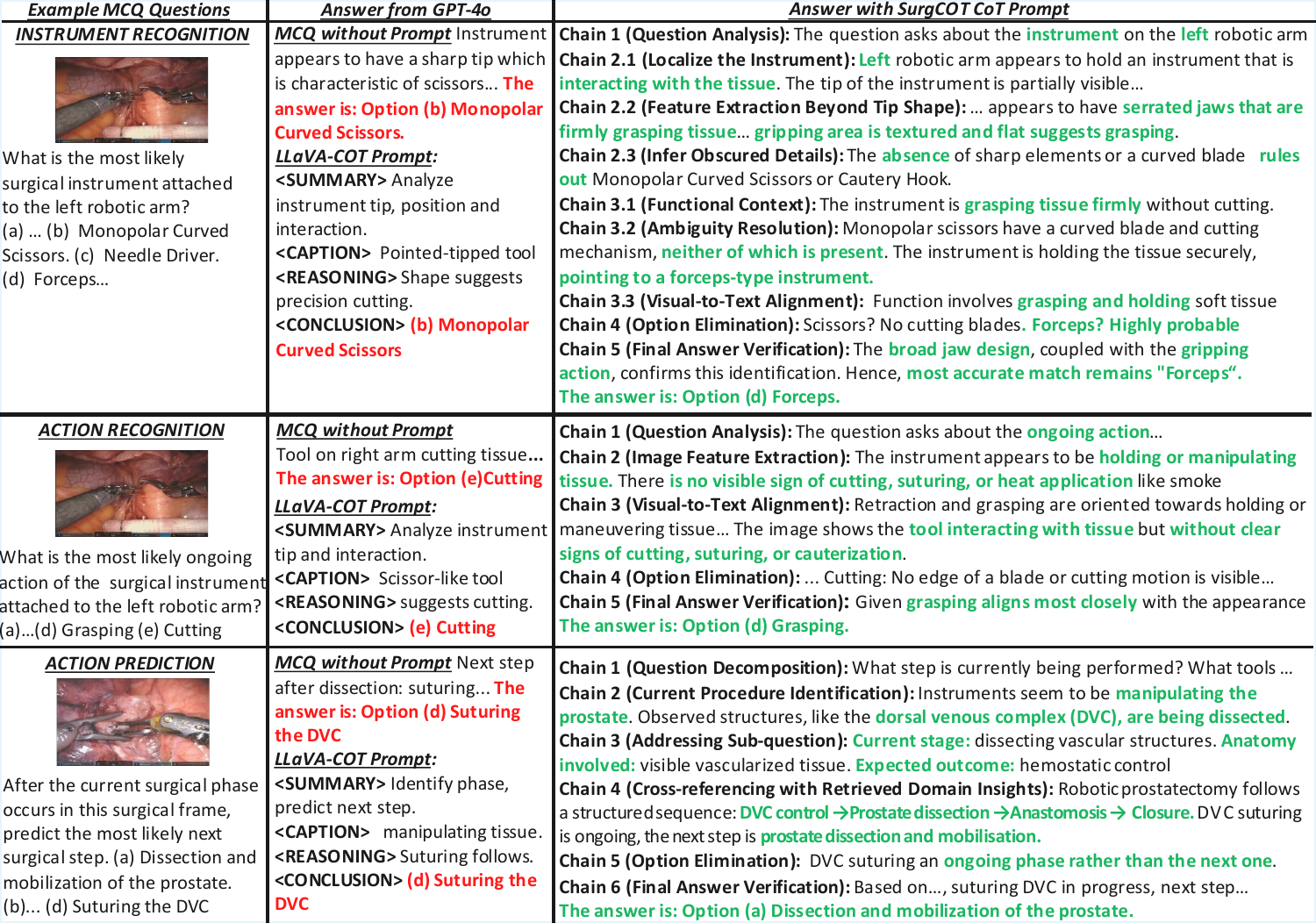}
\caption{Case study for three tasks under different prompts. Red denotes incorrect answers and green denotes correct answers.}
\label{fig:example}
\vspace{-10pt}
\end{figure*}

\setlength{\tabcolsep}{5pt}
\begin{table}[h]
\centering
\caption{Ablation study results (\% mean accuracy) using SurgRAW-GPT4o. RAG denotes Retrieval-Augmented Generation, PD denotes Panel Discussion. Best are bolded.}
\begin{tabular}{cc|c|c|c|c}
\hline
\multicolumn{2}{c|}{Settings} & \multicolumn{4}{c}{Cognitive-Inference Tasks} \\
\hline
CoT & RAG & Act. Pred. & Out. & Pat. & Avg. \\
\hline
& & 
32.98$\pm$5.81 & 
17.92$\pm$7.33 & 
9.44$\pm$5.62 & 
20.11$\pm$2.98 \\
& $\checkmark$ & 
35.68$\pm$4.91 & 
37.83$\pm$5.14 & 
56.04$\pm$4.00 & 
41.32$\pm$2.52 \\
$\checkmark$ & & 
49.66$\pm$4.27 & 
41.51$\pm$4.4 & 
78.16$\pm$3.4 & 
58.97$\pm$2.2 \\
$\checkmark$ & $\checkmark$ & 
\textbf{70.48$\pm$2.81} & 
\textbf{42.87$\pm$3.11} & 
\textbf{96.21$\pm$3.53} & 
\textbf{70.39$\pm$2.16} \\
\hline
\multicolumn{2}{c|}{Settings} & \multicolumn{4}{c}{Visual-Semantic Tasks} \\
\hline
CoT & PD & Act. Rec. & Inst. Rec. & \multicolumn{2}{c}{Avg.} \\
\hline
& & 
37.21$\pm$7.94 & 
43.57$\pm$8.83 & 
\multicolumn{2}{c}{40.39$\pm$3.04} \\
& $\checkmark$ & 
39.33$\pm$6.85 & 
43.57$\pm$8.83 & 
\multicolumn{2}{c}{40.74$\pm$3.01} \\
$\checkmark$ & & 
41.25$\pm$5.61 & 
\textbf{54.13$\pm$1.93} & 
\multicolumn{2}{c}{46.72$\pm$2.81} \\
$\checkmark$ & $\checkmark$ & 
\textbf{44.52$\pm$2.88} & 
\textbf{54.13$\pm$1.93} & 
\multicolumn{2}{c}{\textbf{49.81$\pm$2.31}} \\
\hline
\end{tabular}
\label{tab:ablation1}
\vspace{-10pt}
\end{table}

\vspace{-5pt}
\subsection{Ablation Study}
Table~\ref{tab:ablation1} reveals distinct yet complementary component roles. CoT prompting emerges as the most significant contributor, improving average performance by 38.86\% for CI tasks (20.11\% to 58.97\%) and 6.33\% for VS tasks (40.39\% to 46.72\%). These gains underscore how structured logical inference addresses complex surgical vision tasks. Within CI tasks, RAG provides complementary improvements, with optimal performance achieved through the RAG-CoT combination with an average accuracy of 70.39\%, suggesting retrieval mechanisms are most effective when integrated with systematic reasoning frameworks. For VS tasks, panel discussion is crucial for action recognition, yielding synergistic effects when implemented with CoT-enabled agents to achieve the best performance, showing the complementary roles of supplementary modules in visually ambiguous scene.

\vspace{-5pt}
\section{Conclusion and Future Works}

We introduced SurgRAW, the first agentic system for zero-shot reasoning across major robotic surgical workflow tasks, supporting post-operative analysis and surgical education. While SurgRAW operates on frame-level inputs, these frames are sampled from continuous surgical videos and interpreted within procedural workflow context. CoT prompting is the main contributor to performance gains, with RAG and panel discussion providing complementary support. SurgRAW surpasses prior methods, setting a new benchmark for robotic surgical video understanding. Future work includes broadening SurgCoTBench with more procedures and exploring temporal reasoning and incorporating surgeon-in-the-loop feedback to enhance reliability and clinical impact.
\vspace{-10pt}








\bibliographystyle{IEEEtran}
\bibliography{ref}

\end{document}